\documentclass[conference]{IEEEtran}
\IEEEoverridecommandlockouts
\usepackage{cite}
\usepackage{amsmath,amssymb,amsfonts}
\usepackage{algorithmic}
\usepackage{graphicx}
\usepackage{textcomp}
\usepackage{stmaryrd}
\usepackage{calc}
\usepackage{subcaption}
\usepackage{xcolor}
\def\BibTeX{{\rm B\kern-.05em{\sc i\kern-.025em b}\kern-.08em
    T\kern-.1667em\lower.7ex\hbox{E}\kern-.125emX}}
\begin{document}

\title{Autonomous learning and chaining of motor primitives using the Free Energy Principle
\thanks{This work was funded by the Cergy-Paris University Foundation (Facebook grant) and partially by Labex MME-DII, France (ANR11-LBX-0023-01).}
}

\author{\IEEEauthorblockN{Louis Annabi}
\IEEEauthorblockA{\textit{ETIS UMR 8051}\\
\textit{CY University, ENSEA, CNRS}\\
F-95000, Cergy-Pontoise, France \\
louis.annabi@ensea.fr}
\and
\IEEEauthorblockN{Alexandre Pitti}
\IEEEauthorblockA{\textit{ETIS UMR 8051}\\
\textit{CY University, ENSEA, CNRS}\\
F-95000, Cergy-Pontoise, France \\
alexandre.pitti@ensea.fr}   
\and
\IEEEauthorblockN{Mathias Quoy}
\IEEEauthorblockA{\textit{ETIS UMR 8051}\\
\textit{CY University, ENSEA, CNRS}\\
F-95000, Cergy-Pontoise, France \\
mathias.quoy@ensea.fr}
}

\maketitle

\begin{abstract}
In this article, we apply the Free-Energy Principle to the question of motor primitives learning. An echo-state network is used to generate motor trajectories. We combine this network with a perception module and a controller that can influence its dynamics.

This new compound network permits the autonomous learning of a repertoire of motor trajectories. To evaluate the repertoires built with our method, we exploit them in a handwriting task where primitives are chained to produce long-range sequences.

\end{abstract}


\section{Introduction}


We consider the problem of building a repertoire of motor primitives from an open-ended, task agnostic interaction with the environment. We suggest that a suitable repertoire of motor primitives should enable the agent to reach a set of states that covers best its state space. Based on this hypothesis, we train an agent to learn a discrete representation of its state space, as well as motor primitives driving the agent in the learned discrete states. In a fully observable environment, a clustering algorithm such as Kohonen self-organising maps~\cite{Kohonen1982} applied to the agent's sensory observations make it possible to learn a set of discrete states that covers well the agent's state space. Using this set of discrete states as goals, an agent can learn policies that drive it towards those goals, thus building for itself a repertoire of motor primitives. Our main contribution is to address this twofold learning problem in terms of free energy minimisation.



The Free Energy Principle~\cite{Friston2006} (FEP) suggests that the computing mechanisms in the brain accounting for perception, action and learning can be explained as a process of minimisation of an upper bound on surprise called free energy. On the one hand, FEP applied to perception~\cite{Friston2008} translates the inference on the causes of sensory observations into a gradient descent on free energy, and aligns nicely with the predictive coding~\cite{Rao1999} and Bayesian brain hypotheses~\cite{Dayan1995TheHM}. On the other hand, FEP applied to action, or active inference~\cite{Friston2009a,Friston2016}, can explain motor control and decision making as an optimisation of free energy constrained by prior beliefs. In this work, we present a variational formulation of our problem that allows us to translate motor primitives learning into a free energy minimisation problem.


In previous works, we applied the principle of free energy minimisation in a spiking recurrent neural network for the generation of long range sequences~\cite{Pitti2017b} but not associated to sensorimotor control. The presented model was able to generate long range sequences minimising free energy functions corresponding to several random goals. We used a randomly connected recurrent neural network in order to generate trajectories, and combined it with a second population of neurons in charge of driving its activation into directions minimising the distance towards a randomly sampled goal.

Using randomly connected recurrent neural networks to generate sequences is at the core of reservoir computing (RC) techniques~\cite{Verstraeten2007, Lukosevicius2009}. In particular, there is work using RC for the generation of motor trajectories, see for instance~\cite{Namikawa2010, Mannella2015}. In the RC framework, inputs are mapped to a high-dimensional space by a recurrent neural network called reservoir, and decoded by an output layer. The reservoir weights are fixed and the readout weights are regressed, usually with gradient descent. In~\cite{Pitti2017b}, we proposed to alter the learning problem by fixing the readout weights to random values as well, and by optimising the input of the reservoir network instead.



In this article, we propose to combine the ideas developed in our previous work with a perception module in order to learn a repertoire of motor primitives. We train and evaluate our model in an environment designed for handwriting, where the agent controls a 2 degrees of freedom arm. The agent randomly explores its environment by drawing random trajectories on the canvas. By clustering its visual observations of the resulting trajectories, the agent learns a set of prototype observations which it will sample as goals to achieve during its future drawing episodes. This learning method is interesting from a developmental point of view since it implements goal-babbling. Developmental psychology tells us that learning sensorimotor contingencies~\cite{o'regan2001} plays a key role in the development of young infants. In ~\cite{Jacquey2019}, the authors present a review about sensorimotor contingencies in the fields of developmental psychology and developmental robotics, in which they propose a very general model on how a learning agent should organise its exploration of the environment to develop its sensorimotor skills. They suggest that the agent should continuously sample goals from its state space and practice achieving these goals. The work we propose in this article aligns nicely with their suggestion, as our agent randomly samples goals from a discrete state space, and optimises the motor sequences leading to these discrete states.

In the following we will first present the model, then the results on motor primitives learning. In a third part, we will evaluate the learned repertoires on a task of motor primitive chaining to draw complex trajectories.

\section{Methods}
\label{sec:methods}

\subsection{Model for motor primitives learning}

\begin{figure}[h!]
    \centering
    \def\svgwidth{0.9\columnwidth}
    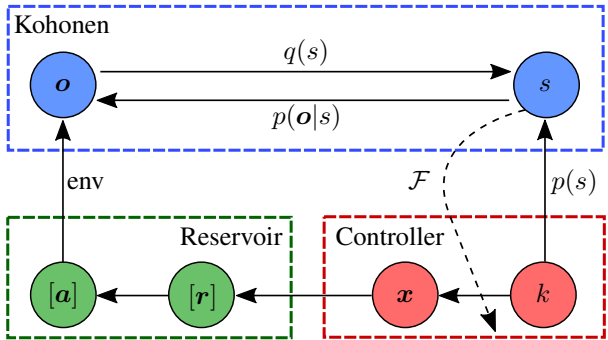
    \caption{Model for motor primitives learning. $\boldsymbol{x}$ denotes the activation signal. $[\boldsymbol{r}]$ denotes the sequence of activations $\{\boldsymbol{r}_t\}_{t=1..T}$ of the reservoir network. $[\boldsymbol{a}]$ denotes the sequence of atomic motor commands $\{\boldsymbol{a}_t\}_{t=1..T}$ decoded from the reservoir dynamics. $\boldsymbol{o}$ denotes visual observation. $s$ denotes the hidden state. $k$ denotes the primitive index on which depend the activation signal $\boldsymbol{x}$ and the prior probability over hidden state $p(s)$. $\mathcal{F}$ denotes the free energy, it is used as an optimisation signal for the controller.}
    \label{fig:model_1}
\end{figure}

Our neural network architecture, represented in figure \ref{fig:model_1}, can be segmented into three substructures. On the sensory pathway (top), a Kohonen map is used to cluster the observations. On the motor pathway (bottom), a reservoir network and a controller are used to model the generation and optimisation of motor sequences.

The activation signal $\boldsymbol{x}$ stimulates the random recurrent neural network (RNN), that exhibits a self-sustained activity $\boldsymbol{r}$ during $T$ time steps. $T$ atomic motor commands $\boldsymbol{a}$ are readout from the $T$ activations of the recurrent network. The environment provides an observation $\boldsymbol{o}$ after being impacted by the sequence of actions. This observation is then categorised in $s \in S = \{s_i\}_{i<n}$ by a Kohonen network.

\subsubsection{Reservoir network}

Motor primitives are sequences of atomic motor commands read out from the dynamics of the reservoir network. To optimise these sequences according to some criteria, one can either optimise the readout weights, the dynamics of the reservoir or control its activity (our case). Updating the recurrent weights of the RNN can have the undesirable effect of limiting its dynamics. Thus in RC, the focus is usually put on the learning of appropriate readout weights. However we showed in a previous work~\cite{Pitti2017b} that it is possible to control the RNN dynamics by optimising its initial state. In this article this is the strategy we use in order to learn motor primitives.

Our implementation of the reservoir network is based on an existing model for handwriting trajectory generation using RC~\cite{Laje2013}, using the following equations :
\begin{align}
    \boldsymbol{u}(t) &= (1 - \frac{1}{\tau}) \cdot \boldsymbol{u}(t-1) + \frac{1}{\tau} (\boldsymbol{W_r} \cdot \boldsymbol{r}(t-1)) \\
    \boldsymbol{r}(t) &= \tanh(\boldsymbol{u}(t)) \\
    \boldsymbol{a}(t) &= \tanh(\boldsymbol{W_o} \cdot \boldsymbol{r}(t)) 
\end{align}
where $\boldsymbol{u}$ and $\boldsymbol{r}$ denote the network activation respectively before and after application of the non-linearity. We denote by $\tau$ the time constant of the network.

For the recurrent network to have a self-sustained activity, we referred to the weights initialisation used in~\cite{Laje2013}. The recurrent weights matrix $\boldsymbol{W_r}$ is made sparse with each coefficient having a probability $p_r$ of being non-null. When non-null, the coefficients are sampled from a normal distribution $\mathcal{N}(0, \frac{\sigma_r^2}{n_r})$ with a variance scaled according to the network size $n_r$. The readout weights $\boldsymbol{W_o}$ are sampled from a normal distribution $\mathcal{N}(0, \sigma_o^2)$.

\subsubsection{Kohonen map}

The Kohonen map~\cite{Kohonen1982} takes as input a 64x64 gray scale image. Each filter learned by the Kohonen map has to correspond to a distinct motor primitive. Since we expected to learn motor primitives corresponding mainly to movement orientations, we used a Kohonen map topology with only one cyclic dimension. We also ran experiments with 2-d and 3-d topologies with and without cyclic dimensions. We chose to stick with the 1-d cyclic topology because it presented a fast learning and a balanced use of all the learned filters. Here are the equations of the Kohonen network:
\begin{align}
    i_w(t) &= \operatornamewithlimits{argmin}_i(\|\boldsymbol{W_k}[i, :](t) - \boldsymbol{o}(t)\|_2^2) \\
    \boldsymbol{W_k}(t+1) &= (1 - \lambda_k) \cdot \boldsymbol{W_k}(t) + \lambda_k \cdot N(i_w(t)) \odot \boldsymbol{o}(t)
\end{align}
where $\boldsymbol{W_k}$ denotes the Kohonen weights, each row $\boldsymbol{W_k}[i, :]$ corresponding to the filter associated with neuron of index $i$. $i_w$ denotes the winner neuron
index, i.e. the index of the Kohonen neuron whose associated filter is closest to the input stimulus $\boldsymbol{o}$. 
The neighbourhood function $N(i_w(t))$ depends on the chosen topology. It is maximum in $i_w(t)$ and decreases exponentially according to the distance with regard to the winner neuron index $i_w(t)$. This exponential decay is parameterised by a neighbourhood width $\sigma_k^2$. The operator $\odot$ denotes the element-wise product.

\subsubsection{Free energy derivations}

Our goal is to learn the right activation signals $\boldsymbol{x}$ that stimulate the random recurrent network in a way that leads to desired categorisations of the observation $\boldsymbol{o}$ by the Kohonen network. Let $n$ be the number of motor primitives that we want to learn. We set the size of the Kohonen network as well as the number of activation signals to learn to $n$. For the trajectory of index $k$, we want to learn the stimulus $\boldsymbol{x_k^*}$ that better activates the corresponding category in the Kohonen network (i.e. such that $p(\boldsymbol{o}|s=s_k) \approx 1$). We can observe that the optimal activation signal $\boldsymbol{x_k^*}$ depends on the weights of the Kohonen map. Because of this dependence, it would be easier to learn and fix the Kohonen map before optimisation of the controller. However, it is more realistic from a developmental point of view to train these two structures at the same time. For this reason, we learn both networks in parallel but control their learning parameters over time, to be able to favor the learning of one network compared to the other.

We use free-energy minimisation as the strategy to train the controller. What follows is a formalisation of our model using a variational approach:

\begin{itemize}
    \item $p(s)$ is the prior probability over states. Here we propose using a softmax, parameterised by $\beta > 0$, around the index $k$ of the current primitive.
    \begin{equation}
        p(s=s_i) = \frac{\exp(-\beta \cdot |k - i|)}{\sum_j{\exp(-\beta \cdot |k - j|)}}
    \end{equation}
    
    \item $p(\boldsymbol{o}|s)$ is the state observation mapping. The observations are images of size $d$. For simplicity, we make the approximation of considering all pixel values as independent. We choose to use Bernoulli distributions for all pixel values $o_{l<d} \in \{0, 1\}$. Since all pixel values are considered independent, the probability distribution over the whole observation can be factorised as:
    \begin{equation}
        p(\boldsymbol{o}|s=s_i) = \prod_{l<d}{\boldsymbol{W_k}[i, l]^{o_l} \cdot (1-\boldsymbol{W_k}[i, l])^{1 - o_l}}
    \end{equation}
    where $\boldsymbol{W_k}[i, l]$ is the value of pixel $l$ of the filter $i$ of the Kohonen map.
    
    \item $q(s)$ is the approximate posterior probability over states knowing the observation $\boldsymbol{o}$. Here we define $q(s)$ to be the one-hot distribution over states such that $q(s=s_i) = \delta_{i_w, i}$, where $i_w$ is the index of the Kohonen neuron with the highest activation (i.e. whose filter is the closest to the observation): $i_w = \text{argmin}_j(\|\boldsymbol{W_k}[j, :] - \boldsymbol{o}\|_2^2)$.
\end{itemize}

We can now derive the free-energy computations using this model:
\begin{align}
    \centering
    \mathcal{F}_1(\boldsymbol{o}) &= \text{KL}(q(s)||p(s)) - \sum_{i<n}{q(s_i) \log(p(\boldsymbol{o}|s_i))} \\
    &= \sum_{i<n}{q(s_i)\log \frac{q(s_i)}{p(s_i)}} - \sum_{i<n}{q(s_i) \log(p(\boldsymbol{o}|s_i))} \\
    &= - \log(p(s_{i_w})) - \log(p(\boldsymbol{o}|s_{i_w}))
\end{align}
The first term of the free energy in eq. (8) is a quantity called complexity. It scores how complex the approximate posterior is compared to the prior. It decreases when $q(s)$ and $p(s)$ are close. In our case, it is minimal when $i_w = k$, meaning that the category chosen by the Kohonen map is the one with the highest prior probability. Minimising complexity thus induces the network to generate trajectories that activate the right Kohonen category.

The second term is the opposite of the quantity called accuracy. Accuracy measures how good the approximate posterior probability $q(s)$ is at predicting the observation $\boldsymbol{o}$. Here, it increases when the Kohonen filter of the winner neuron is close to the observation. Maximising accuracy induces the network to generate trajectories that are as close as possible to one of the Kohonen filter. For simplicity, we will call this quantity inaccuracy instead of opposite of accuracy.

Summing those two quantities, minimising free energy would result in observations that are close to one of the Kohonen filter, and in this Kohonen filter being the one with the highest prior probability. 

\subsubsection{Optimisation method}

Our optimisation problem is the following. For each primitive $k$, we want to find an activation signal $\boldsymbol{x_k}$ that generates an observation $\boldsymbol{o}$ resulting in a low free energy $\mathcal{F}_1(\boldsymbol{o})$. To use gradient based methods, we would need to have a diiferentiable model of how the activation signals $\boldsymbol{x_k}$ impacts the resulting free energy $\mathcal{F}_1(\boldsymbol{o})$. Since we do not have a model of how the environment produces observations, the whole $\boldsymbol{x}\rightarrow \boldsymbol{r} \rightarrow \boldsymbol{a} \rightarrow \boldsymbol{o} \rightarrow \mathcal{F}_1(\boldsymbol{o})$ chain is not differentiable. To solve our problem, we instead use a random search optimisation method, detailed by the following algorithm.

\begin{algorithmic}
\STATE \COMMENT {Random initialisation of the controller}
\FOR {$k<n$}
    \STATE $\boldsymbol{x}_k \gets \mathcal{N}(0, 1)$
\ENDFOR
\STATE \COMMENT {Training}
\FOR {$e < E$}
    \STATE $k \gets \mathcal{U}(n)$
    \STATE $\boldsymbol{\delta x} \sim \mathcal{N}(0, \sigma^2(e))$
    \STATE $u_+ \gets \boldsymbol{x_k} + \boldsymbol{\delta x}$
    \STATE $u_- \gets \boldsymbol{x_k} - \boldsymbol{\delta x}$
    \STATE $[\boldsymbol{a}_1, \dots, \boldsymbol{a}_T]_+ \gets $ simulate\_action($u_+$) 
    \STATE $[\boldsymbol{a}_1, \dots, \boldsymbol{a}_T]_- \gets $ simulate\_action($u_-$) 
    \STATE $\boldsymbol{o}_+ \gets $ env($[\boldsymbol{a}_1, \dots, \boldsymbol{a}_T]_+$) 
    \STATE $\boldsymbol{o}_- \gets $ env($[\boldsymbol{a}_1, \dots, \boldsymbol{a}_T]_-$) 
    \STATE $i_{w,+} \gets $ simulate\_kohonen($\boldsymbol{o}_+$)
    \STATE $i_{w,-} \gets $ simulate\_kohonen($\boldsymbol{o}_-$)
    \STATE $f_+ \gets  $ free\_energy($i_{w,+}, \boldsymbol{o}_+$)
    \STATE $f_- \gets  $ free\_energy($i_{w,-}, \boldsymbol{o}_-$)
    \STATE $\boldsymbol{x}_k \gets \boldsymbol{x}_k - \lambda \cdot (f_+ - f_-) \cdot\boldsymbol{\delta x} $
\ENDFOR 
\end{algorithmic}

The parameter $e$ in the search standard deviation $\sigma^2(e)$ indicates that this coefficient can depend on the training episode $e$. The "simulate\_action" function used in the code above corresponds to the iterative application of equations (1), (2), (3) for the duration $T$ of the motor primitives. The "env" function corresponds to the generation of an observation by the environment after being impacted by the sequence of actions. This computation is performed by the environment and unknown to the agent. The "simulate\_kohonen" function corresponds to the application of equations (4) and (5). The "free\_energy" function corresponds to the application of equation (10).

\subsection{Experimental setup}

\subsubsection{Environment}
\begin{figure}[h!]
    \centering
    \includegraphics[width=0.40\textwidth]{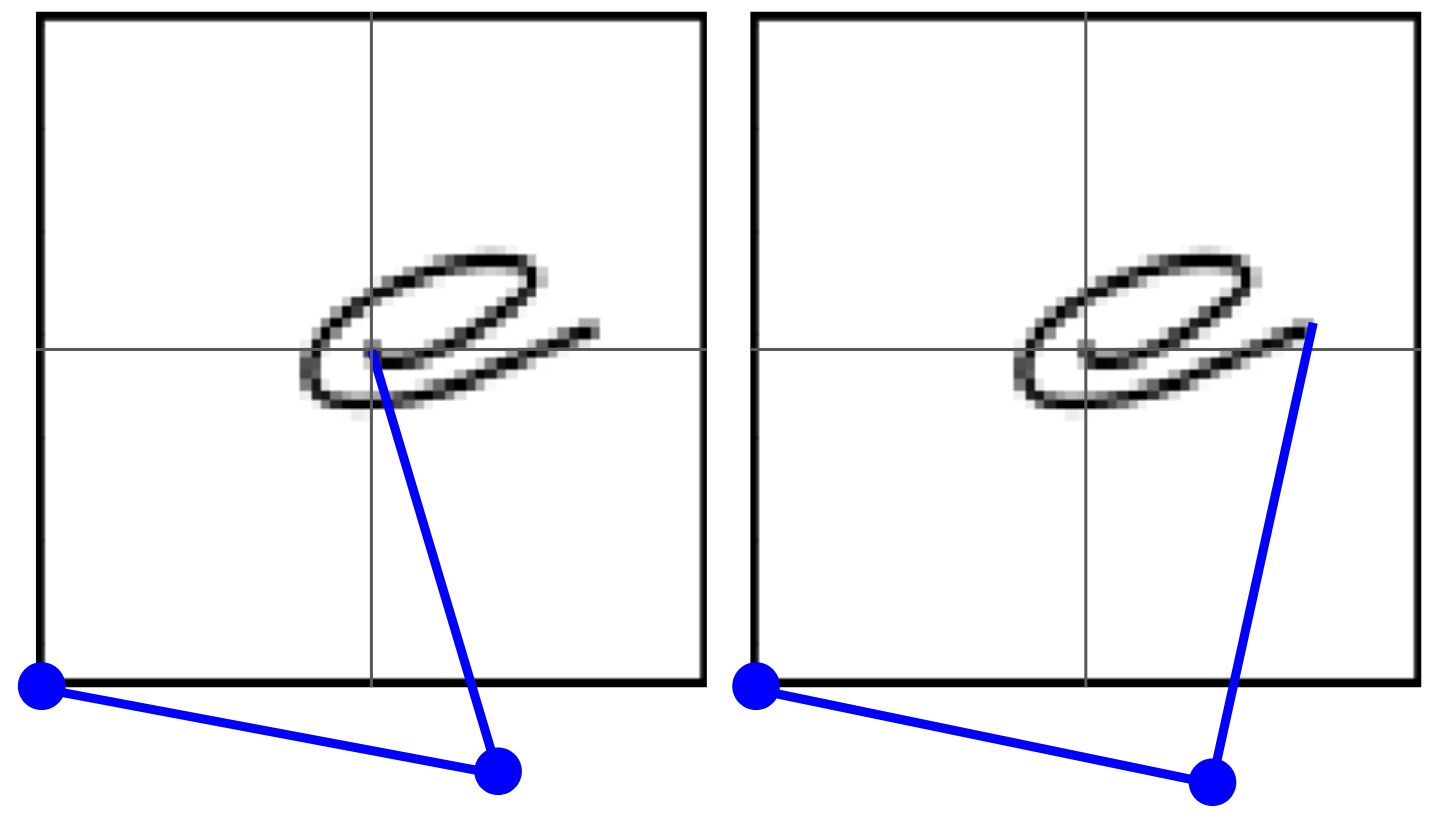}
    \caption{Initial (left) and final (right) arm position for a trajectory taken from a data set of handwriting trajectories.}
    \label{fig:robot_arm}
\end{figure}

The environment is an initially blank canvas on which the agent can draw. The initial position of the pen is at the center of the canvas. The agent can act on the environment via 2D actions. The actions are the angle velocities of a 2 degrees of freedom arm as represented in figure \ref{fig:robot_arm}.

\subsubsection{Training}

We start from a random initialisation of the Kohonen map. Over time, the Kohonen map self-organises when being presented with the trajectories generated by the random search algorithm. Simultaneously, the random search algorithms learns to generate motor trajectories that lead to the different Kohonen prototype observations.

\begin{figure}[h!]
    \centering
    \includegraphics[width=\columnwidth]{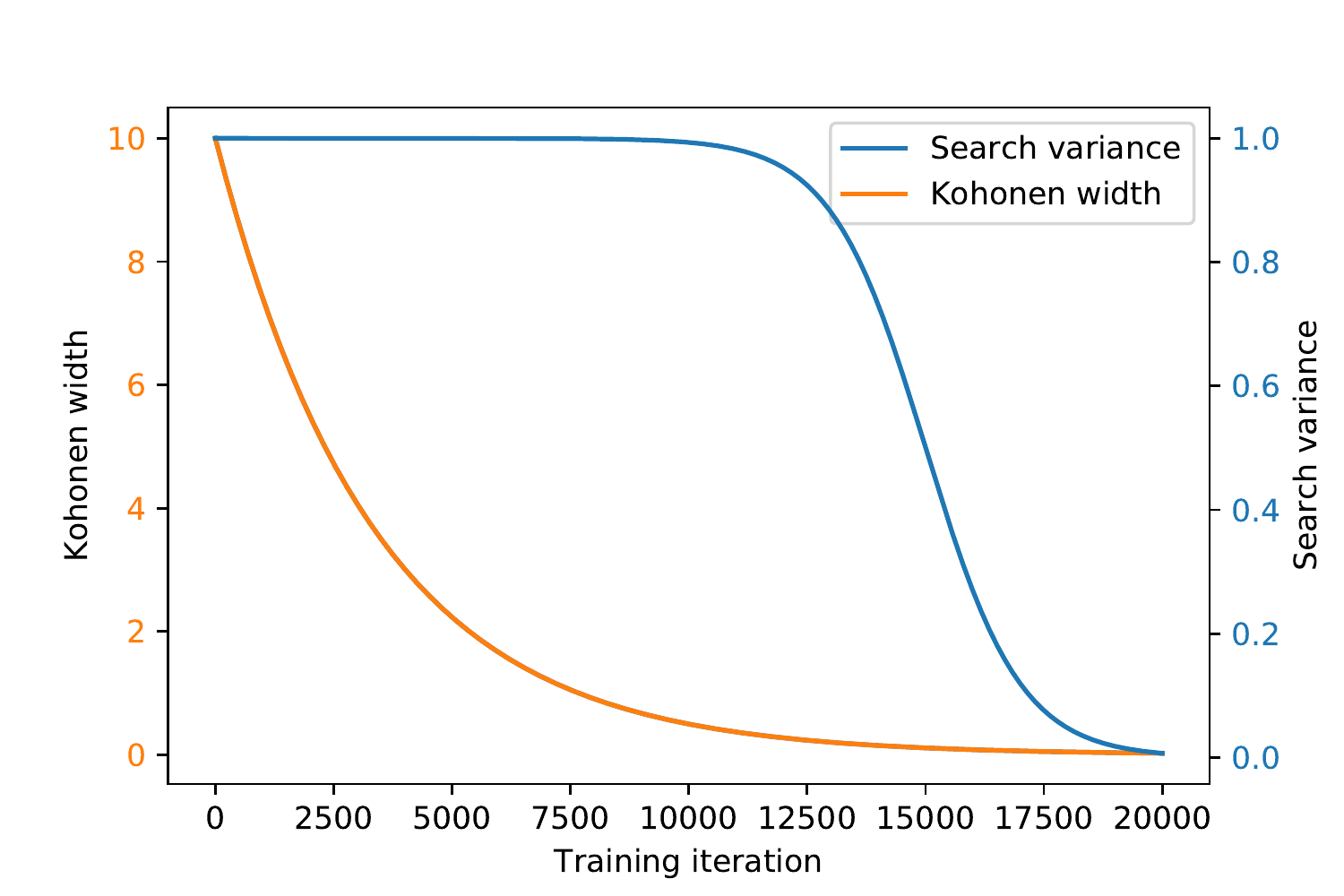}
    \caption{Search variance $\sigma^2(e)$ and Kohonen width $\sigma_k^2(e)$ according to training iteration $e$.}
    \label{fig:hyperparameters}
\end{figure}

Training was performed using the following set of parameters for the different components described in the previous section:

\begin{itemize}
    \item RNN: $n_r=100$, $\tau = 10$, $p_r = 0.1$, $\sigma_r^2 = 1,5$.
    \item Readout layer: $n_o = 2$, $\sigma_o^2 = 1$.
    \item Kohonen map: $n = 50$, $\lambda_k=0.01$, Kohonen width $\sigma_k^2(e)$ varies over time, see figure \ref{fig:hyperparameters}.
    \item Free energy: $\beta \in \{2^{-5}, 2^{-4}, 2^{-3}, 2^{-2}, 2^{-1}, 1, 2, 4, 8, 16\}$.
    \item Random optimisation: $\lambda=0.01$, $\sigma^2(e)$ varies over time, see figure \ref{fig:hyperparameters}.
\end{itemize}

\subsection{Results}

We trained our model for $E=20000$ iterations on $n=50$ primitives. At each iteration, we uniformly sample $k$ from $[ 1, n ]$. We train on the $k$\textsuperscript{th} primitive by adjusting the prior probability as in (6) and optimising $\boldsymbol{x_k}$. On average, each activation signal $\boldsymbol{x_k}$ is trained on $400$ iterations. Figure \ref{fig:hyperparameters} displays the evolution of the random optimisation search variance and of the Kohonen width over the 20000 iterations. We discuss this choice in the light of the following results.

\begin{figure}[h!]
    \centering
    \includegraphics[width=\columnwidth]{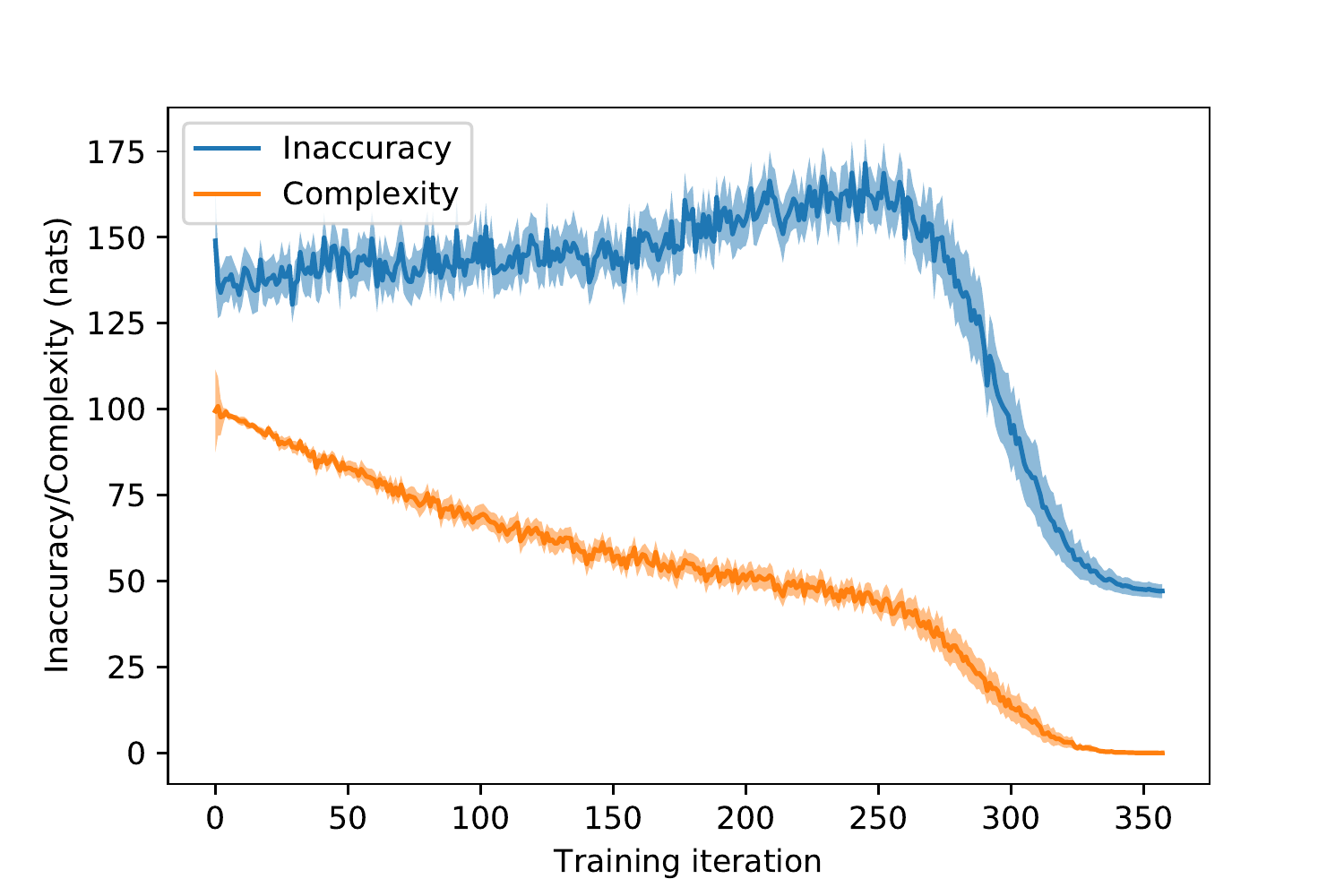}
    \caption{Inaccuracy and complexity averaged on the number of primitives $n=50$, with $\beta=8$. Each primitive has been sampled on at least 360 iterations, and on average on 400 iterations.}
    \label{fig:results_beta_8_angle}
\end{figure}

Figure \ref{fig:results_beta_8_angle} displays the evolution of inaccuracy and complexity during training.

During the first phase, when $e<12500$, the random search has a very high variance. Consequently, the trajectories generated and fed to the Kohonen map are very diverse and this allows the Kohonen map to self-organise. Inaccuracy does not seem to decrease in this early phase. This is because the Kohonen filters, initially very broad, are becoming more precise. The high variance in the random search allows for a diminution of complexity but still generates trajectories that are too noisy to accurately fit the more precise Kohonen filters.

During the second phase, we decrease the variance of the random search. The system can now converge more precisely and this causes a faster decrease of both inaccuracy and complexity.

We can question whether it is necessary for the random search variance to remain high for such a long time, since it slows down learning. We observed that if we reduce the duration of the first phase, the Kohonen does not have the time to self organise and this results in an entangled topology. Because of the complexity term in the free energy computations, the topology of the Kohonen has an influence over the learning. For instance, with an entangled topology, a search direction for $\boldsymbol{x_k}$ that activates a neuron closer to $k$ might not make the actual trajectory closer to the ones recognised by the $k$\textsuperscript{th} Kohonen neuron. In other words, having a proper topology smooths the loss function.

We also notice that the inaccuracy cannot decrease below a certain value. At first, we could think that this is because the optimisation strategy is stuck in a local optimum. However, we obtained the same lower bound on inaccuracy over different training sessions. Since the optimisation strategy relies on random sampling, there is no evident reason to encounter the same local minimum. Our explanation is that this lower bound is imposed by the Kohonen network neighbourhood function. Because the Kohonen width does not reach 0, the Kohonen centroids are still attracting each other and this prevents them from completely fitting the presented observations. In consequence, the filters are always partly mixed with their neighbours and this causes the inaccuracy to plateau at a value that depends on $\sigma_k^2$.

\begin{figure}[h!]
    \centering
    \includegraphics[width=0.48\textwidth]{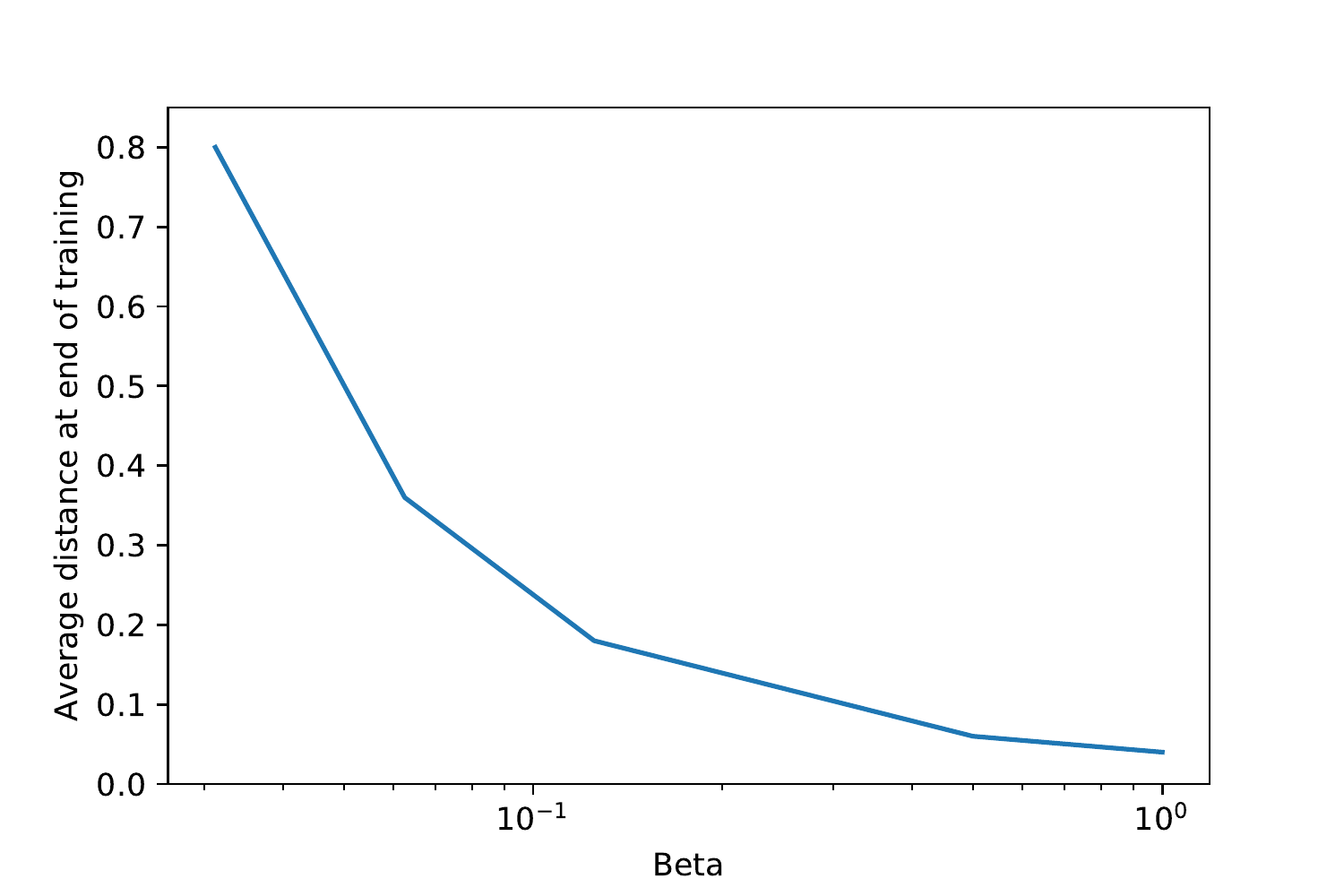}
    \caption{Average distance $|i_w-k|$ between the activated neuron in the Kohonen $i_w$ and the primitive index $k$, according to $\beta$.}
    \label{fig:beta}
\end{figure}

Figure \ref{fig:beta} shows the impact of the parameter $\beta$ over the convergence. Looking at the equations (6) and  (10) we can see that this parameter directly scales the overall complexity. For low values of $\beta$, the random search is more likely to be stuck in local minima of free-energy, when activating a Kohonen neuron closer to $k$ corresponds to an increase in inaccuracy that exceeds the decrease in complexity. We measured the average distance between the activated neuron in the Kohonen and the primitive index $k$ at the end of training for $\beta \in \{2^{-5}, 2^{-4}, 2^{-3}, 2^{-2}, 2^{-1},2^{0}\}$. The results, presented in figure \ref{fig:beta}, confirm that the final states obtained with higher values of $\beta$ correspond to a more precise mapping between $i_w$ (winner index of the Kohonen map) and $k$ (state index enforced by the prior probability).

\begin{figure}[h!]
    \centering
    \includegraphics[width=0.48\textwidth]{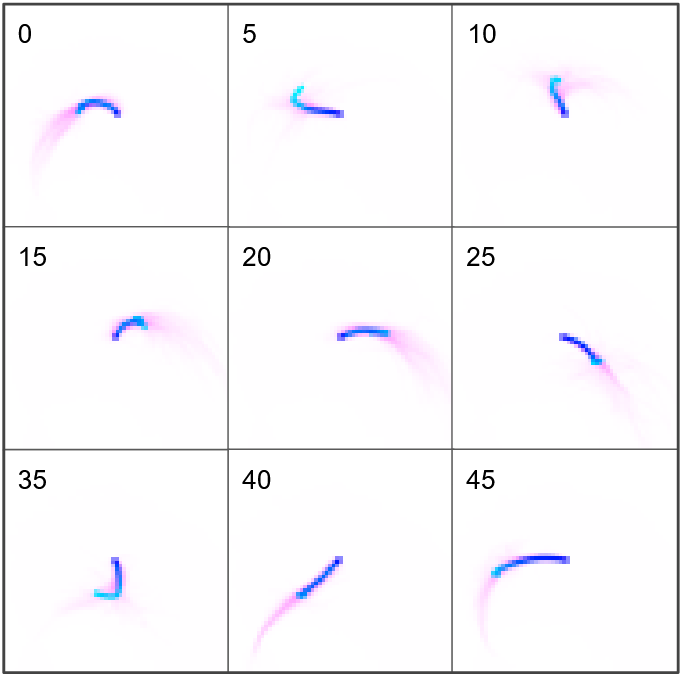}
    \caption{9 of the 50 learned motor primitives and corresponding Kohonen filters: $k \in \{0, 5, 10, 15, 20, 25, 35, 40, 45\}$. The blue component of the image corresponds to the trajectory that is actually being generated by the network for the activation signal $\boldsymbol{x_k}$. The red component of the image corresponds to the Kohonen filter of index $k$.}
    \label{fig:filters_beta_16_angle}
\end{figure}

Figure \ref{fig:filters_beta_16_angle} displays some of the learned motor primitives. The blue component of the image corresponds to the trajectory that is actually being generated by the reservoir network for the activation signal $\boldsymbol{x_k}$. The red component of the
image corresponds to the Kohonen filter of index $k$. 
This figures allows visual confirmation of several points. First, the inaccuracy at the end of training seems to indeed come from the blurriness of the Kohonen filters. Second, the filters and motor primitive trajectories seem to follow a topology: the index of the primitive seems highly correlated with the orientation of the route taken by the arm end effector. Finally, every trajectory seems to be in the center of the corresponding Kohonen filter, which suggests that the minimisation of complexity successfully enforced the mapping between $i_w$ and $k$.

\section{Chaining of motor primitives}

To validate our approach, we still need to show that this repertoire of motor primitives is efficient at constructing more complex movements. To perform this evaluation, we propose to extend the model presented in section \ref{sec:methods}.

First, we define a new perception module meant to classify sensory observations into states corresponding to more complex trajectories. To avoid confusion, we will denote this new hidden state $\sigma$. 

Second, we enable in this revisited model the chaining of motor primitives. In the previous model, each drawing episode corresponds to one motor trajectory of length $T$ being generated. Here in each drawing episode the agent will draw a trajectory corresponding to $M$ motor primitives of length $T$ chained together.

Since we are simply trying to evaluate the primitives learned in the first model, we won't address the training of the network used to classify sensory observations. Instead, our focus will be on the decision making process, i.e. the selection of the primitives $\{k_1, \dots, k_M\}$ to chain in order to reach a certain desired state $\sigma^*$.

\subsection{Model}

\begin{figure}[h!]
    \centering
    \def\svgwidth{0.9\columnwidth}
    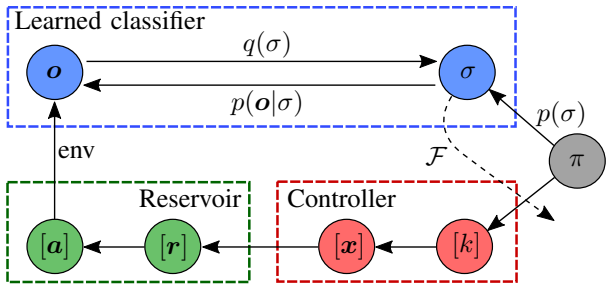
    \caption{Model for motor primitives chaining. $\pi=p(\sigma)$ denotes the prior probability over states $\sigma$. $[k]$ denotes the sequence of motor primitives indices $\{k_m\}_{m=1..M}$. $[\boldsymbol{x}]$ denotes the sequence of corresponding activation signals $\{\boldsymbol{x}_m\}_{m=1..M}$ .$[\boldsymbol{r}]$ denotes the resulting sequence of activations $\{\boldsymbol{r}_{m,t}\}_{m=1..M, t=1..T}$ of the reservoir network. $[\boldsymbol{a}]$ denotes the sequence of atomic motor commands $\{\boldsymbol{a}_{m,t}\}_{m=1..M, t=1..T}$ decoded from the reservoir dynamics. $\boldsymbol{o}$ denotes the visual observation provided by the environment after being modified by the sequence of atomic motor commands. $\sigma$ denotes the hidden state, it is different from the hidden state of the previous section. $\mathcal{F}$ denotes the expected free energy, it is used as an optimisation signal for the choice of sequence of motor primitives indices $\{k_m\}_{m=1..M}$.}
    \label{fig:model_2}
\end{figure}

The model for motor primitives chaining is presented in figure \ref{fig:model_2}.

\subsubsection{New hidden state}

The hidden state $\sigma$ corresponds to new categories representing complex motor trajectories. We used the Character Trajectories Data Set from \cite{Dua2019} composed of approximately 70 trajectories for each letter of the alphabet that can be drawn without lifting the pen. The trajectories are provided as sequences of pen positions. We drew these trajectories using our drawing environment and used the resulting observations to build the new state observation mapping.

\subsubsection{Expected free energy derivations}

The model uses active inference for decision making. It selects actions that minimise a free energy function constrained by prior beliefs over hidden states. According to active inference, constraining the prior probability over states to infer a state distribution that favors a target state $\sigma^*$ will force the agent to perform actions that fulfill this prediction. In this sense, the prior probability over states can be compared to the definition of a reward in reinforcement learning. For instance, a rewarding state would be a state that is more likely under the prior probability over states. 

Here is the formalisation of this model in the variational framework:

\begin{itemize}

    \item Prior probability over states $p(\sigma)$ acts similarly to a reward function in reinforcement learning. The prior beliefs (or prior preferences) over $\sigma$ will be set manually to different values during testing to guide the agent into a desired state.
    
    \item The other probability distribution over states is one that the agent has control on. By choosing one primitive in the learned repertoire, the agent selects one resulting distribution over states $q_k(\sigma)$ modeling how the choice of motor primitive $k$ will influence the state. This probability distribution over hidden states corresponds to the output of the learned classifier when fed with the observation resulting from the application of the $k^{th}$ motor primitive.
    
    \item  As in the primitive learning model presented in section II, the state observation mapping $p(\boldsymbol{o}|\sigma)$ is built using  equation (7). The filters used for each category correspond to the average of the observations belonging to this class obtained from the data set.

\end{itemize}

We can now derive the expected free-energy using this model:
\begin{equation}
    \centering
    \mathbb{E}[\mathcal{F}_2(k)] = \text{KL}(q_k(\sigma)||p(\sigma)) - \sum_{i<n}{q_k(\sigma_i) \mathcal{H}(p(\boldsymbol{o}|\sigma_i))}
\end{equation}
where $\mathcal{H}(p(\boldsymbol{o}|\sigma_i))$ denotes the entropy of the state observation mapping for state $i$.

When the agent has to make a decision about which motor primitive to use, it computes its expected future free energy $\mathbb{E}[\mathcal{F}_2(k)]$ for each possible primitive and selects the primitive with the minimum expected free energy. 

This time again, expected free energy can be segmented as complexity and inaccuracy. Minimising the first term of equation (11) will result in choosing motor primitives that lead to hidden states matching our prior preferences. This can be seen as directly optimising reward. Minimising inaccuracy will result in choosing actions that lead to hidden states with high precision (low entropy) in the state observation mapping. This connects to the drive of surprise avoidance inherent to the free energy principle.

\subsubsection{Action selection}

The following algorithm details the action selection process on a trajectory composed of $M$ motor primitives using free energy minimisation :

\begin{algorithmic}
\STATE $p(\sigma) \gets $ init()
\FOR {$m<M$}
    \FOR{$k<n$}
        \STATE $u \gets \boldsymbol{x}_k$
        \STATE $[\boldsymbol{a}_1, \dots, \boldsymbol{a}_T] \gets $ simulate\_action($u$) 
        \STATE $\boldsymbol{o} \gets $ env\_model($[\boldsymbol{a}_1, \dots, \boldsymbol{a}_T]$)
        \STATE $q_k(\sigma) \gets $ classifier($\boldsymbol{o}$)
        \STATE $f_k \gets $ expected\_free\_energy($q_k(\sigma), p(\sigma), \boldsymbol{o}$)
    \ENDFOR
    \STATE $k_m^* \gets $ arg$\min_k(f_k)$
    \STATE $u^* \gets \boldsymbol{x}_{k_m^*}$
    \STATE $[\boldsymbol{a}_1, \dots, \boldsymbol{a}_T] \gets $ simulate\_action($u$) 
    \STATE env($[\boldsymbol{a}_1, \dots, \boldsymbol{a}_T]$)
\ENDFOR

The function "env\_model" corresponds to a learned forward model that allows estimating the future observations for any sequence of actions. In our experiments, we simply simulated the actions and rewound the environment, which (in a deterministic environment) corresponds to having a perfect forward model (env = env\_model).
The function "classifier" corresponds to the classification of the observation by a learned classifier. It outputs a probability distribution over hidden states $\sigma$.
Finally, the function "expected\_free\_energy" corresponds to the application of eq. (11).

\end{algorithmic}

\subsection{Results}

\begin{figure}[h!]
    \centering
    \begin{subfigure}[t]{0.32\columnwidth}
        \centering
        \includegraphics[width=\textwidth]{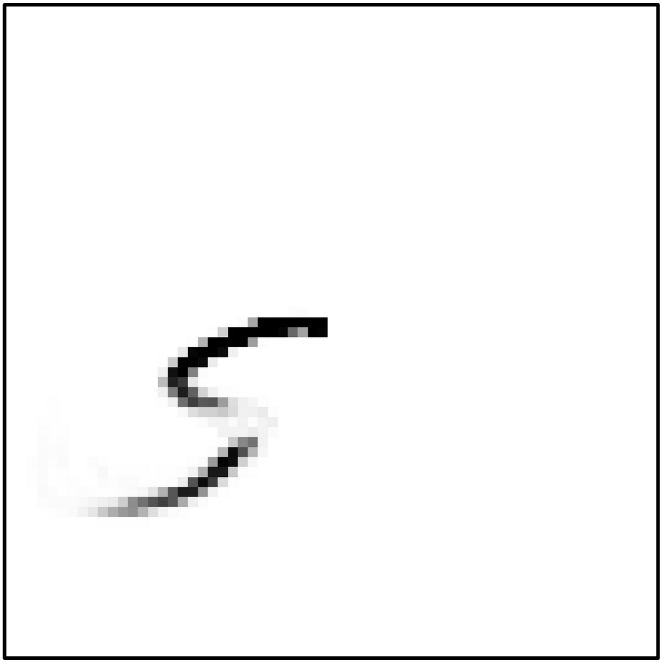}
        \caption{}
        \label{fig:s_a}
    \end{subfigure}%
    ~ 
    \begin{subfigure}[t]{0.32\columnwidth}
        \centering
        \includegraphics[width=1\textwidth]{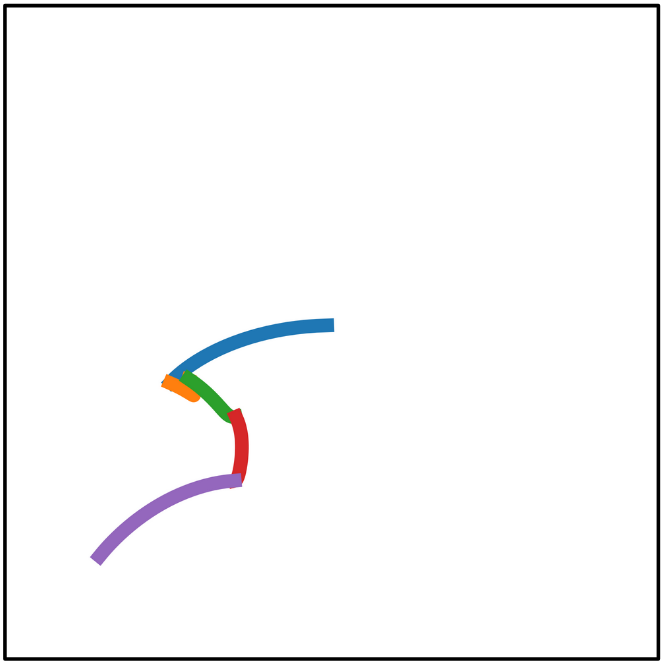}
        \caption{}
        \label{fig:s_b}
    \end{subfigure}%
    ~ 
    \begin{subfigure}[t]{0.32\columnwidth}
        \centering
        \includegraphics[width=1\textwidth]{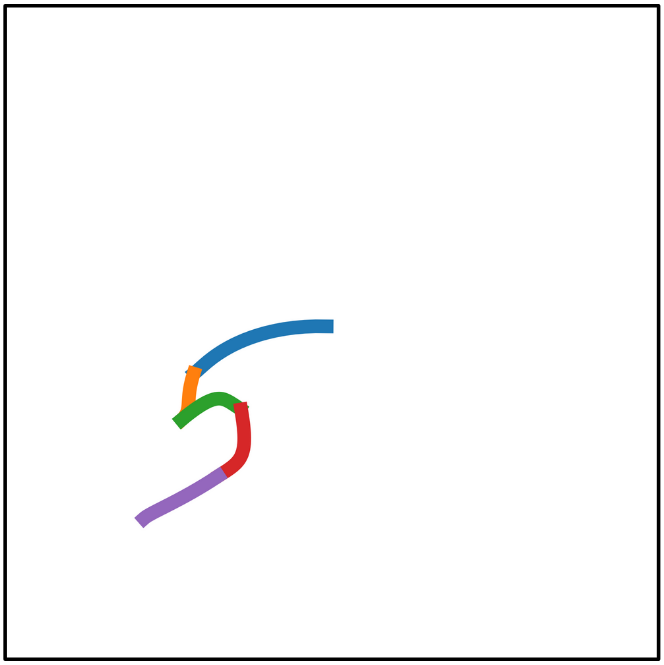}
        \caption{}
        \label{fig:s_c}
    \end{subfigure}
    \caption{Example of filter and produced trajectories. Left: Filter for the category 's'. Middle and right: trajectories produced by chaining five motor primitives belonging to two different learned repertoires.}
    \label{fig:s}
\end{figure}

In our tests, the hidden state $\sigma$ was build using five classes corresponding to the letters 'c', 'h', 'i', 's', 'r'. We chose these letters because the trajectories inside each category were relatively close and this allowed for filters of low entropy.

Figure \ref{fig:s} displays one filter of the learned classifier and two trajectories generated by chaining five motor primitives from two different primitive repertoires learned with our model. The trajectories were obtained by setting the prior preferences to 0.96 for the category 's' and 0.01 for the four other categories.

To verify that our model learns a valuable repertoire of motor primitives, we compare the quality of the constructed complex trajectories (as in \ref{fig:s_b} and \ref{fig:s_c}) with our model and with random repertoires.

Random repertoires are built using the same RNN and readout layer initialisations. They differ from the learned repertoires in the fact that we do not optimise the activation signals $\boldsymbol{x_k}$ of the reservoir. The initial states of the reservoir used to generate the primitives are taken as $\boldsymbol{x}_k \sim \mathcal{N}(0,1)$. In other words, they are equivalent to the repertoires our model would provide before learning.

\begin{figure}[h!]
    \centering
    \includegraphics[width=0.48\textwidth]{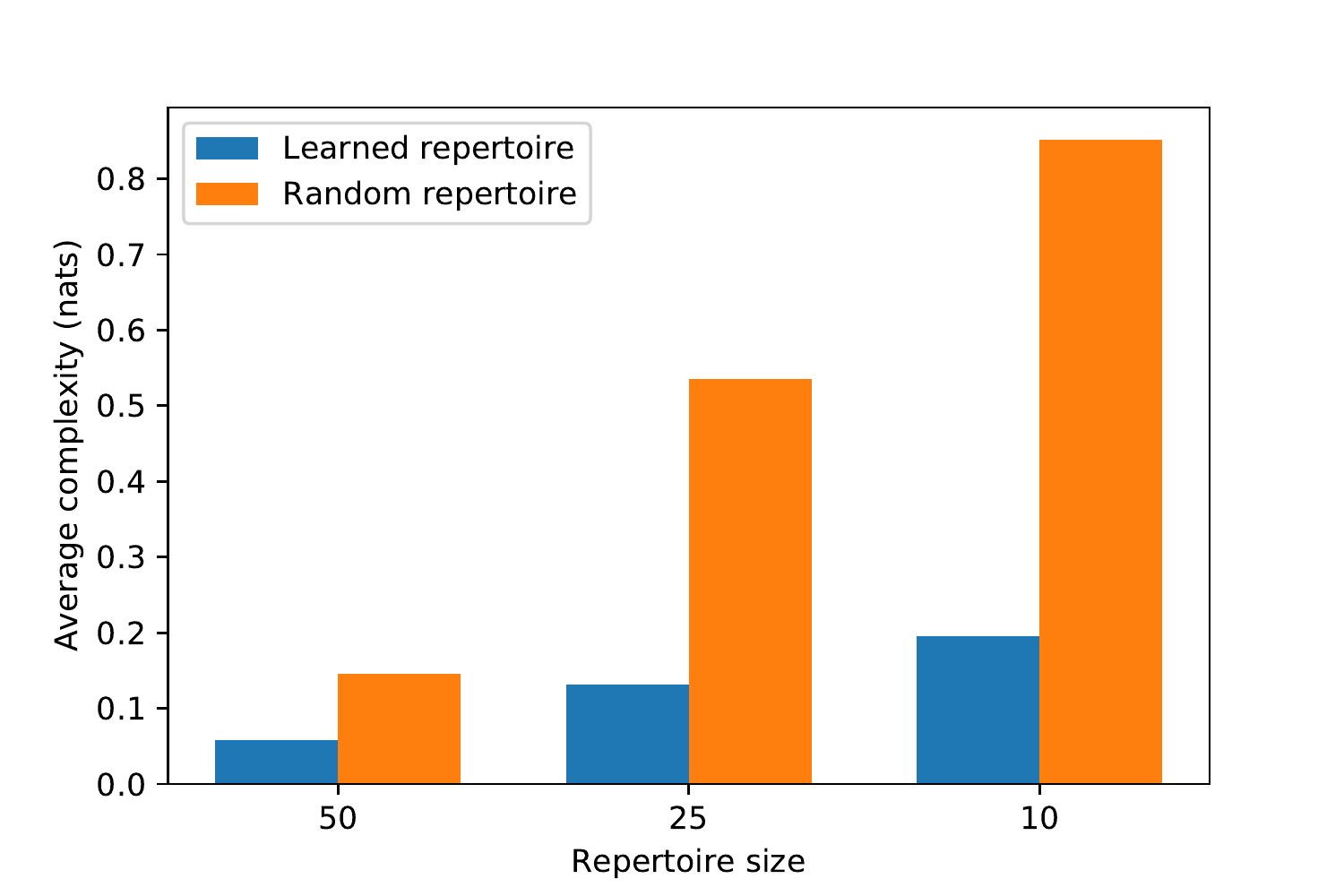}
    \caption{Average complexity for learned and random repertoires of different sizes.}
    \label{fig:performance_bar_chart}
\end{figure}

Figure \ref{fig:performance_bar_chart} displays the average complexities measured at the end of the episode. For each episode, we set the prior preferences of one of the letter category to 0.96 and the others to 0.01. The values are averaged over the different letters and for 5 different repertoires, learned or random.

The complexity scores how close the recognition probability $q(\sigma)$ (provided by the learned classifier) is to the prior preferences $p(\sigma)$ and thus constitutes a suitable indicator for comparison. For low complexities, the constructed images are close to the filter, as in \ref{fig:s}.

We observe that the average complexity tends to be lower for repertoires of larger sizes, independently of the type of repertoire. Having a larger repertoire of primitives indeed should be an asset in order to reconstruct more complex trajectories. For every repertoire size, we measure a lower complexity with repertoires learned using the model described in section II. 

\section{Conclusion}

The results displayed in section III show that our model is able to learn repertoires of motor primitives that are efficient at building more complex trajectories when combined. 

To further validate our approach, it would be interesting to compare our results with other strategies for motor primitive learning. On the one hand, there is existing work in developmental robotics prescribing guidelines to build repertoires of motor primitives \cite{Jacquey2019}, \cite{ugur2012}, but they don't provide neural network implementation to be used for comparison.

On the other hand, the option discovery literature in hierarchical reinforcement learning provides practical methods to build repertoires of relevant options. Options were introduced in~\cite{SUTTON1999181} as a candidate solution to address the issue of temporal scaling in reinforcement learning. An option is defined as a temporally extended action, and thus is conceptually similar to a motor primitive. It would be interesting to measure how our approach compares with current state of the art techniques for unsupervised option learning such as~\cite{Eysenbach2018}.

\section*{Acknowledgment}

This work was funded by the Cergy-Paris University Foundation (Facebook grant) and Labex MME-DII, France (ANR11-LBX-0023-01).

\bibliographystyle{IEEEtran}
\bibliography{references}

\end{document}